\documentclass{article}
\usepackage{color,soul}
\usepackage{spconf,amsmath,graphicx,amsfonts}
\usepackage[normalem]{ulem}
\usepackage[utf8]{inputenc}
\usepackage{xcolor}
  
\newcommand\blfootnote[1]{%
  \begingroup
  \renewcommand\thefootnote{}\footnote{#1}%
  \addtocounter{footnote}{-1}%
  \endgroup
}
\usepackage{afterpage}

\newcommand\blankpage{
    \null
    \thispagestyle{empty}
    \addtocounter{page}{-1}
    \newpage
    }

\title{Fourier Transform of Percoll Gradients Boosts CNN Classification of Hereditary Hemolytic Anemias}

\name{\begin{tabular}{c}Ario Sadafi$^{1,3\star}$ \qquad Lucía María Moya Sans$^{1,2\star}$ \qquad Asya Makhro$^{4}$ \qquad  Leonid Livshits$^{4}$\\  Nassir Navab$^{3,5}$ \qquad Anna Bogdanova$^{4\star\star}$ \qquad Shadi Albarqouni$^{3,6\star\star}$ \qquad Carsten Marr$^{1\star\star}$ \end{tabular}}

\address{$^1$ Institute of Computational Biology, Helmholtz Center Munich,\\ German Research Center for Environmental Health, Munich, Germany \\
$^2$ Universidad Carlos III de Madrid, Madrid, Spain \\
$^3$ Computer Aided Medical Procedures, Technical University of Munich, Munich, Germany \\
$^4$ Red Blood Cell Research Group, Institute of Veterinary Physiology, Vetsuisse Faculty and the Zurich \\Center for Integrative Human Physiology, University of Zurich, Zurich, Switzerland \\
$^5$ Computer Aided Medical Procedures, Johns Hopkins University, Baltimore, USA\\
$^6$ Helmholtz AI, Helmholtz Center Munich, Neuherberg, Germany}

\begin{document}
%\ninept
%
% \topskip0pt
% \vspace*{\fill}
\onecolumn

\hspace{0pt}
\vfill

{\Huge IEEE Copyright Notice}
\\[0.7in]
{\large © 2021 IEEE. Personal use of this material is permitted. Permission from IEEE must be obtained for all other uses, in any current or future media, including reprinting/ republishing this material for advertising or promotional purposes, creating new collective works, for resale or redistribution to servers or lists, or reuse of any copyrighted component of this work in other works.}
\\[.3in]
\textbf{\large
Pre-print of article that will appear at the 2021 IEEE International Symposium on Biomedical Imaging (ISBI 2021), April 13-16 2021}
% \vspace*{\fill}
\hspace{0pt}
\vfill

\blankpage{}
\twocolumn

\maketitle
\blfootnote{$^{\star}$ shared first authorship}
\blfootnote{$^{\star\star}$ shared senior authorship}

\begin{abstract}
Hereditary hemolytic anemias are genetic disorders that affect the shape and density of red blood cells. Genetic tests currently used to diagnose such anemias are expensive and unavailable in the majority of clinical labs. Here, we propose a method for identifying hereditary hemolytic anemias based on a standard biochemistry method, called Percoll gradient, obtained by centrifuging a patient's blood. Our hybrid approach consists on using spatial data-driven features, extracted with a convolutional neural network and spectral handcrafted features obtained from fast Fourier transform. We compare late and early feature fusion with AlexNet and VGG16 architectures. AlexNet with late fusion of spectral features performs better compared to other approaches. We achieved an average F1-score of 88\% on different classes suggesting the possibility of diagnosing of hereditary hemolytic anemias from Percoll gradients. Finally, we utilize Grad-CAM to explore the spatial features used for classification.
\end{abstract}
\begin{keywords}
Image Classification, Deep Learning, Red Blood Cells, Percoll Density Gradients.
\end{keywords}

\section{Introduction}
Hereditary hemolytic anemias are a group of disorders caused by genetic mutations that affect shape and density of red blood cells. Red blood cells have various channels and pumps helping them to release ions such as calcium and potassium in and out of the cell. An intact membrane structure allows them to expand or shrink according to the environment, leading to higher or lower densities. Many hereditary hemolytic disorders are directly affecting either the membrane, the pumps, or the channels, thus leading to abnormal shape and density of cells. The variance in red blood cell density has recently been suggested to serve as a marker of severity of hereditary spherocytosis \cite{huisjes2020density} and sickle cell disease \cite{mahkro2020pilot}.

\begin{figure*}[t]
    \centering
    \includegraphics[width=\textwidth, trim=0 10.3cm 0 0,clip]{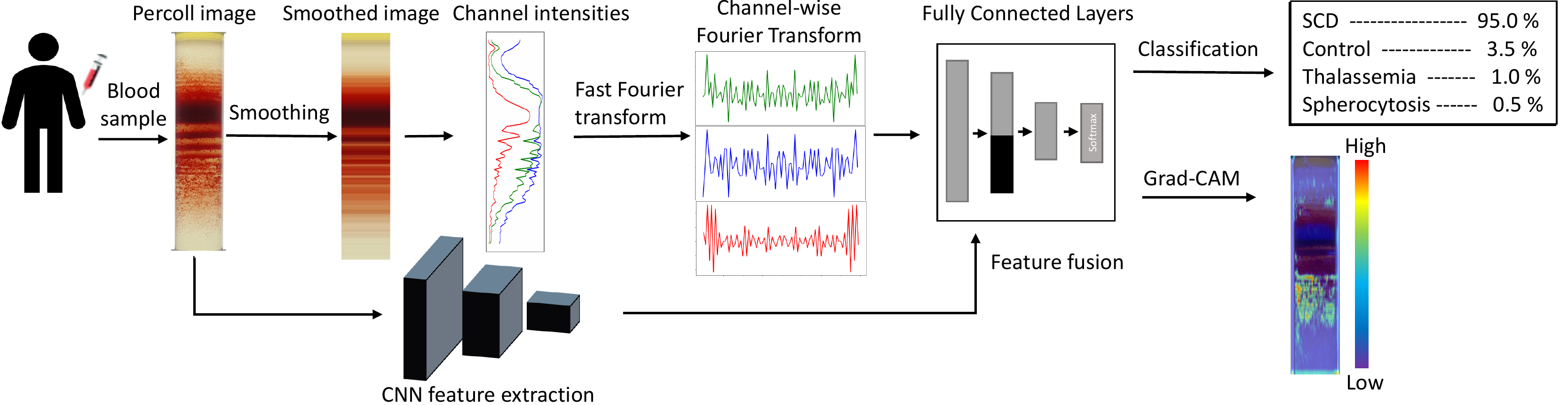}
    \caption{Overview of the proposed method. Blood samples are obtained from patients and centrifuged to obtain Percoll gradient images. Two different feature extraction approaches are employed for classification: CNN feature extraction and fast Fourier transform applied on the smoothed images. After classification, Grad-CAM is used to highlight important regions of the image.}
    \label{figoverview}
\end{figure*}

Percoll is a standard density gradient medium for cell and particle separation in biochemistry. It has low viscosity, low osmolarity, and is not toxic and thus an ideal tool to investigate red blood cell density distribution. Percoll density gradients \cite{rbcfrontiers} form bands with different thicknesses. These bands might hold important information about the patient blood cells' aggregation tendency and consistency (Fig. \ref{figoverview}). 

In many rural areas, access to medical facilities is limited, and the need for developing AI solution for affordable healthcare is highly desirable. 
We were wondering if we can identify a hereditary hemolytic anemia from a Percoll gradient, a simple and cheap experimental approach as compared to a genetic test which is expensive and still scarce in clinical labs. To that end, we collected Percoll gradients from patients suffering from sickle cell disease, thalassemia, and spherocytosis along with healthy controls and tried to classify the samples with a deep neural network as a proof of concept. 

Deep learning approaches are being widely developed for use in the medical domain~\cite{litjens2017survey} covering a wide spectrum of medical imaging ranging from CT, MRI scans to microscopic imaging~\cite{matek2019human}, histopathology~\cite{peng2019multi}, and cancer diagnosis~\cite{bejnordi2017diagnostic}.     
Some of these methods are concentrated on red blood cells and their relevant diseases. For instance, Manescu et al. \cite{manescu2020weakly} are suggesting a weakly supervised method for the diagnosis of malaria and sickle cell disease. In our recent work \cite{sadafi2020attention} we suggested a multiple instance learning approach for the classification of red blood cell disorders. Such methods can be highly useful for diagnosis provided that a microscope is available on site. 

\textbf{Contributions} In this work, we are proposing a method based on convolutional neural networks and fast Fourier transform to classify Percoll gradients for patient diagnosis in hereditary hemolytic anemias. To the best of our knowledge, this is the first work trying to detect hemolytic anemias by looking at Percoll gradients. We investigate two different combinations of spatial features with the handcrafted features from the Fourier transform to find the best fusion scheme that boosts the accuracy of the method. The first results are highly promising and encourage further investigations on this topic.

\section{Method}
Our proposed method (Fig.~\ref{figoverview}) combines both the spatial data-driven features from convolutional neural networks (CNN) and the spectral handcrafted features using Fourier transform to perform Percoll gradients classification. Formally, for a given dataset $\mathcal{D}=\{(I_1, c_1), \dots, (I_N, c_N)\}$, where $I_i \in \mathbb{R}^{H \times W \times 3}$ is the Percoll image, and $c_i \in \{sickle, thalassemia, spherocytosis, healthy\}$ is the corresponding class label, our objective is to build a model $f(\cdot)$ that predicts the class label  $\hat{c}_q$ for a given query Percoll image $\hat{c}_q = f(I_q; \Theta)$, where $\Theta$ is the model parameters. 

\subsection{Spatial data-driven features}
Typical CNNs used for image classification consist of two sections: (i) A convolutional part for feature extraction and (ii) fully connected layers for classification. In our method, we opt for the convolutional part to extract spatial data-driven features, denoted CNN features, 
\begin{equation}
    h_{\mathrm{cnn}} = f_{\mathrm{cnn}}(I, \theta), 
\end{equation}
where $\theta \subset \Theta$ are the convolutional parameters.

\subsection{Spectral handcrafted features}
%change here to add motivation for the fourier transform addition
We assume that most of RBC information such as viscoelasticity, amount of hemoglobin, density and aggregability, is retained in the band pattern of the Percoll gradients. This pattern can be interpreted as a signal and analyzed.
% in which the principal frequencies correspond to a group of RBCs with specific properties and that is why Fourier transform was included as a supplement for the feature extraction in the CNN.}
To extract spatial features using the Fourier transform, a smoothed version $\overline{I} \in \mathbb{R}^{\frac{H}{n}\times 3}$ 
% (which was thought as a "signal without noise") 
is obtained from the input image $I$ (Fig.~\ref{figoverview}) by averaging a neighborhood of size $n\times n$ along the y axis of the image as follows

\begin{equation}
    \overline{I}_k^r =\frac{1}{n^2} \sum_{i=n(k-1)}^{n.k} \sum_{j = l -\lfloor \frac{n}{2} \rfloor}^{ l + \lfloor\frac{n}{2}\rfloor} I_{(i, j, r)}, 
\end{equation}
where $k\in[1,\frac{H}{n}]$, $l = \lfloor\frac{W}{2}\rfloor$, and $W$ and $H$ are width and height of the input image, respectively, and $r$ is the selected channel of the image. 
The smoothed image helps eliminating unwanted noise in the bands. Intensities of each channel are sampled separately. 

\textbf{Fourier transform} We use the Fast Fourier Transform (FFT) algorithm \cite{cooley1965algorithm} to compute the discrete Fourier transform of the sampled sequences $I_{\mathrm{fft}} = \mathrm{fft}(\overline{I})$. 
We normalized the Fourier transform for every channel as follows

\begin{equation}
   h_{\mathrm{fft}} = \frac{I_{FFT}}{(H/2n)} = \frac{2n \dot I_{FFT}}{H}.
\end{equation}

\subsection{Feature fusion}
\begin{figure}[ht]
    \centering
    \includegraphics[width=0.45\textwidth,page=2, trim=0 11.1cm 19.5cm 0,clip]{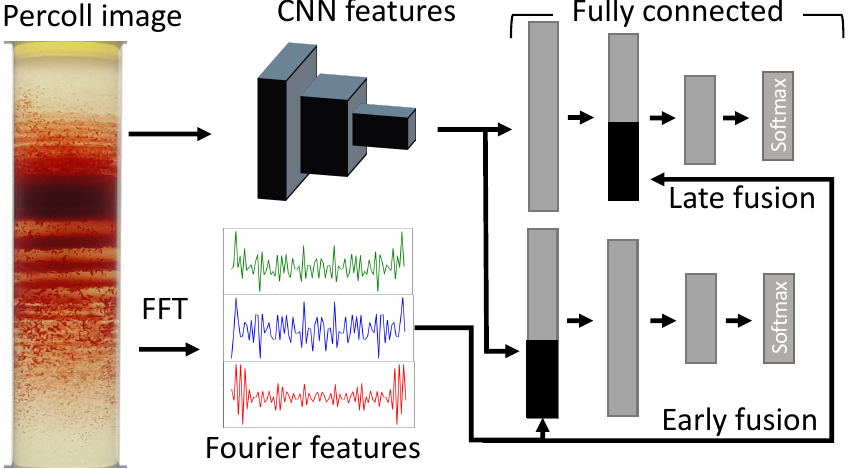}
    \caption{Two approaches of early fusion and late fusion of the spectral features are used and compared. In early fusion fast Fourier transform (FFT) features are fused at the first fully connected layer while in late fusion features are fused at the second layer.}
    \label{figEFLF}
\end{figure}

The classifier section of the CNN consists of three fully connected layers. Two different approaches of early fusion and late fusion were designed to incorporate the FFT features (Fig.~\ref{figEFLF}). In early fusion (EF), Fourier features $h_{fft}$ are injected into the first fully connected layer while in late fusion features are injected to the second fully connected layer.

\begin{figure*}[ht!]
    \centering
    \includegraphics[width=0.90\textwidth,page=3, trim=0 10.5cm 0 0,clip]{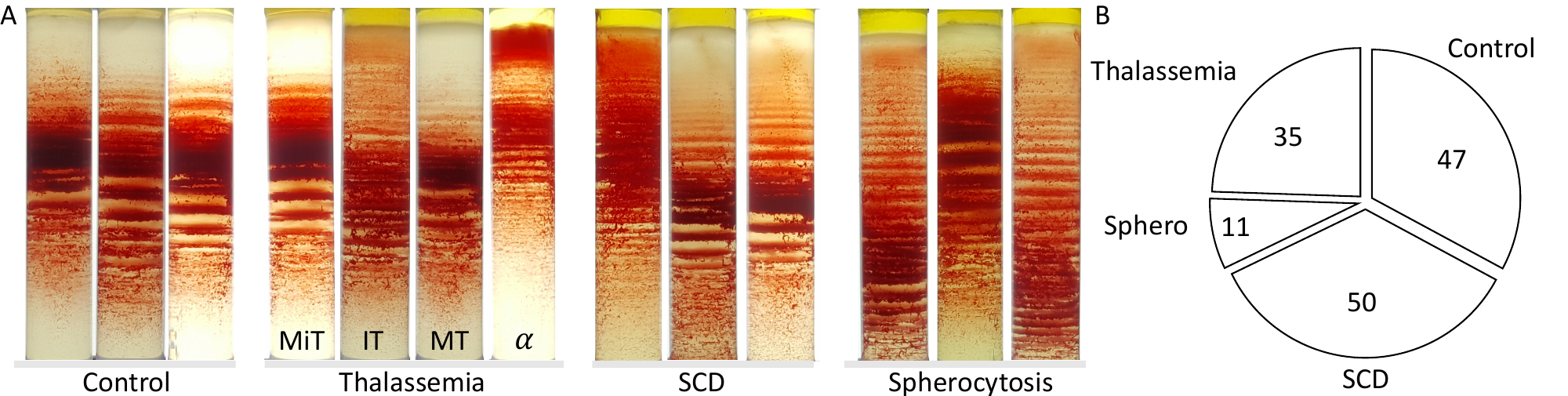}
    \caption{(A) Exemplary images of the dataset belonging to different classes. Minor (MiT), intermediate (IT), major (MT) and alpha($\alpha$) thalassemia sub-types are displayed as well. (B) Pie chart showing the class distribution.}
    \label{figdataset}
\end{figure*}
More formally, the input image $I_i$ belonging to class $c_i$, CNN features $h_{\mathrm{cnn}}$ are fused with corresponding Fourier transform features $h_{\mathrm{fft}}$ of the smoothed images, and passed to the classifier $f_{\mathrm{cls}}(\cdot)$ to minimize the following objective function,   

\begin{equation}
    \mathcal{L}_{cls}(\theta,\phi) = \mathrm{CE}(c_i, \hat{c}_i), 
\end{equation}
where $\hat{c}_i = f_{\mathrm{cls}}(h_{\mathrm{cnn}} \odot h_{\mathrm{fft}}; \phi)$ is the predicted class, CE is the cross entropy loss and $\phi \subset \Theta$ is the parameters of the fully connected part of the CNN.

\section{Experiments and results}

\subsection{Dataset}

The dataset consists of 143 patients collected from the Pediatric Hematology Unit and the Laboratory Division of the Emek Medical Center in Afula and are carefully processed. The test tubes look identical and images are obtained by placing the tubes in holders set up for this purpose with white background and lighting to minimize the batch effect. The ground-truth of samples comes from the genetic test of the patients. 
 The 143 patients comprise 50 affected with sickle cell disease, 35 with thalassemia (5 alpha, 30 beta of which 4 of minor, 5 of transfusion independent, and 21 of transfusion dependent), 11 with spherocytosis and 47 controls (Fig.~\ref{figdataset}B).
%Include here the sentence:The amount of people with these rare diseases is small so the database would be increased in future, when more subjects of study are found (?)

\subsection{Implementation details}
The proposed method consists of three components: Convolutional layers used for feature extraction, fast Fourier transform, and fully connected classifier.

\textit{Convolutional layers}: We decided to test with two different standard networks: AlexNet~\cite{krizhevsky2012imagenet} and VGG16~\cite{simonyan2014very}. With 2D adaptive average pooling the extracted tensors of the networks were changed into feature vectors of 9216 and 25088 for AlexNet and VGG16 respectively. All networks were trained using stochastic gradient descend with a learning rate of 0.0001 and momentum of 0.9.
%Change here, added the optimizer

\textit{Fast Fourier transform}: The standard implementation of FFT in SciPy package was used. The sampled sequences each have 100 values for every channel obtained from $5\times 5$ neighborhoods ($n=5$). Having RGB images as input, after FFT analysis 300 values were yield as extracted features.

\textit{Fully connected classification}: Three fully connected layers with Rectified Linear Unit (ReLU) activation functions were designed. To avoid over-fitting, two dropout layers with a rate of $0.5$ also were used during the training.

\textit{Training}: We opt for 3-fold cross validation. Images in training folds are augmented with vertical flipping, cropping, translation, and random noise. The models parameters were optimized using stochastic gradient descend with a learning rate of $0.0001$ and a momentum of $0.9$. AlexNet and VGG16 models were trained for 30, and 20 epochs, respectively. Further information can be found in our repository under https://github.com/marrlab/percollFFT.

\textit{Evaluation metrics}: 
Accuracy, weighted F1 score, area under ROC and Precision recall curve as well as the confusion matrix were calculated using SciPy package.

\subsection{Results}
We repeated each experiment five times and averaged each metric and report mean ± standard deviation.
We are comparing three different experiments: late and early Fourier features fusion with models without any feature fusion as the baseline.
Table \ref{results} shows the results of the three experiments for AlexNet and VGG16 models. 
\begin{table}[t]
\caption{Comparison between AlexNet and VGG16 when trained individually and when accompanied by early fusion (EF) and late fusion (LF) of FFT features. All experiments are run 5 times to report mean and standard deviation.}
\label{results}
\resizebox{0.48\textwidth}{!}{  
\begin{tabular}{l|c|c|c}
    Method & Accuracy & F1 Score & AU ROC\\
    \hline\hline
    AlexNet & $0.86\pm0.02$ & $0.86\pm0.02$ & $0.9722\pm0.0036$\\
    + EF & $0.85\pm0.01$ & $0.84\pm0.02$ & $0.9622\pm0.0037$\\
    + LF & $\mathbf{0.88\pm0.01}$ & $\mathbf{0.88\pm0.01}$ & $\mathbf{0.9770\pm0.0027}$\\
    \hline
    VGG16 & $0.83\pm0.01$ & $0.83\pm0.01$ & $0.9540\pm0.0041$\\
    + EF & $0.84\pm0.01$ & $0.84\pm0.01$ & $0.9639\pm0.0033$\\
    + LF & $0.85\pm0.01$ & $0.85\pm0.01$ & $0.9751\pm0.0030$\\
\end{tabular}
}
\end{table}

\begin{figure}[ht]
    \centering
    \includegraphics[width=0.48\textwidth,page=6, trim=0.3cm 3.65cm 3.8cm 0.1cm,clip]{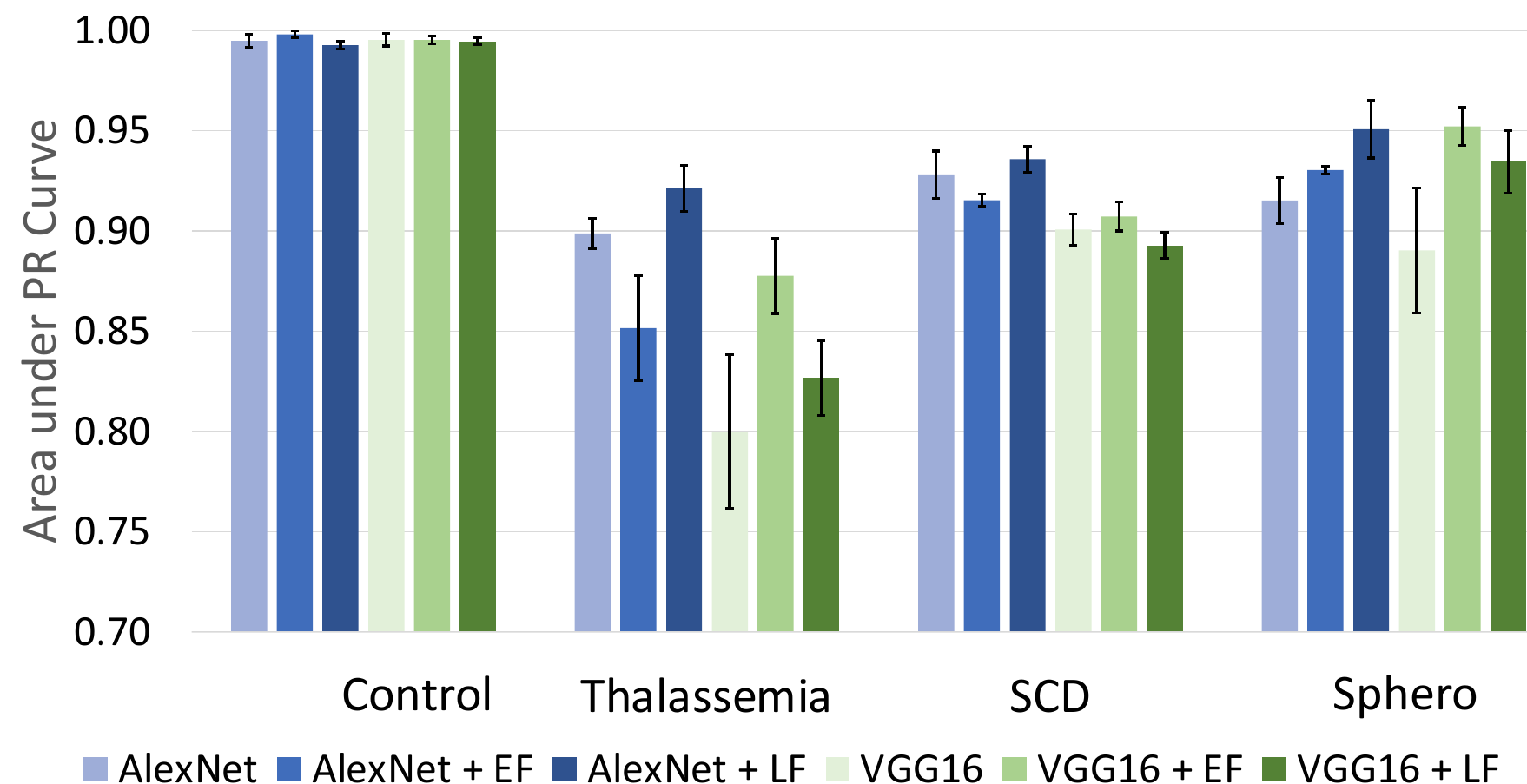}
    \caption{Area under precision recall curve for all experiments and every class is demonstrated with mean and standard deviation of five independent runs.}
    \label{figPRCurve}
\end{figure}

All of the methods are performing better with late fusion of Fourier features. Due to the nature of our particular problem, fast Fourier transform generates valuable features that once fused with typical CNNs can lean to robust classification of Percoll gradient images. Figure \ref{figPRCurve} demonstrates the area under precision recall curve for all of the models across different classes. 
Surprisingly, AlexNet is outperforming VGG16 by a margin of 3\% which can be attributed to the fact that VGG16 has twice the number of parameters of AlexNet and might not be suitable for this task.
Note that thalassemia class consists of three sub-classes based on the severity of the disease. In case of major thalassemia, patients have to receive biweekly transfusions, and donor blood may change the consistency of their blood hence the lower performance on Fig. \ref{figPRCurve}.

\subsection{Grad-CAM}
\begin{figure}[t]
    \includegraphics[width=0.47\textwidth,page=7,trim=0 5.5cm 12.2cm 0cm,clip]{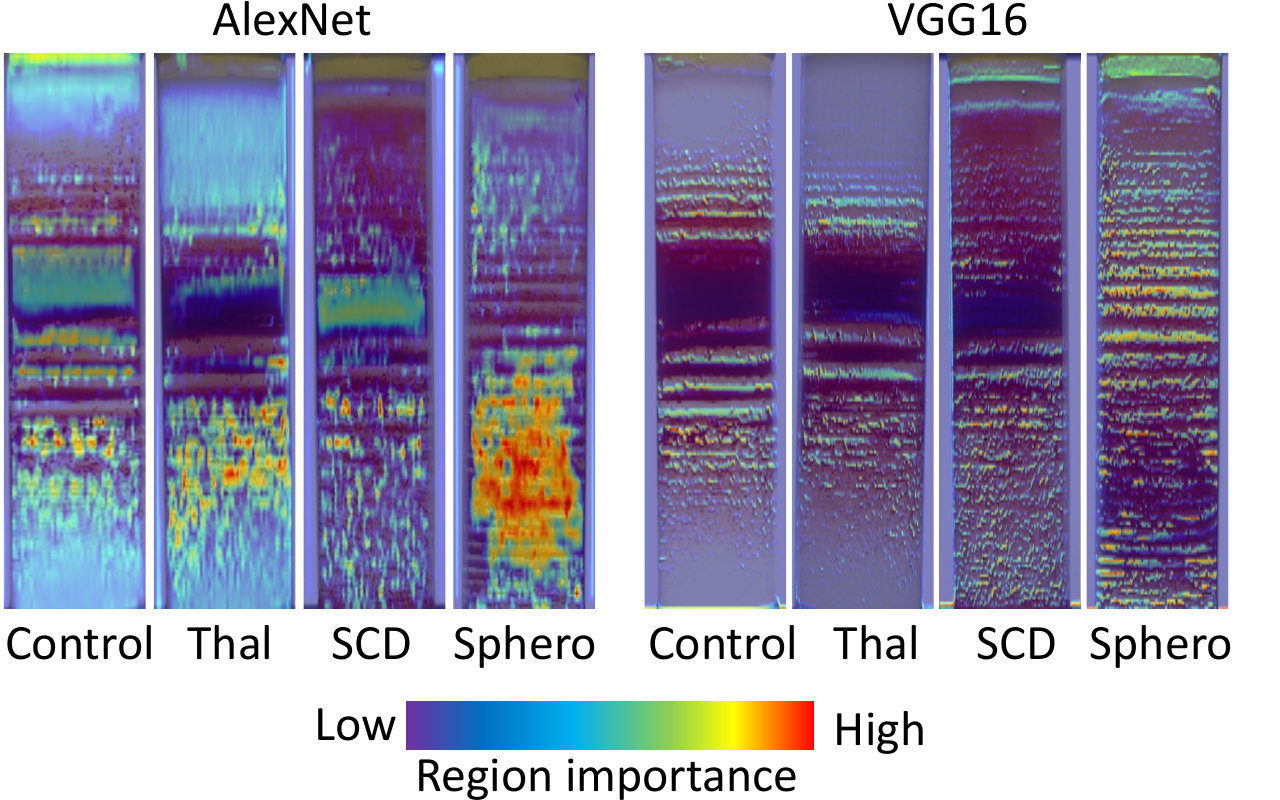}

    \caption{Grad-CAM on correctly classified samples. While AlexNet focuses more on the consistency of the blood, VGG16 favors the bands and density distribution.} 
    \label{figGradcam}
\end{figure}

When dealing with a small dataset, one of the biggest concerns is that the trained models may overfit on irrelevant features. Moreover in medical application explainability is crucial. We decided to use Gradient-weighted Class Activation Mapping (Grad-CAM) \cite{selvaraju2017grad}, as a simple method to make sure our model is actually taking the Percoll gradient bands into account. Grad-CAM is a method that enables visual explanation for decision of CNN-based models by monitoring the gradients of the output logits all the way to the final convolutional layer to highlight the important regions by generating a coarse localization map.
We carried out Grad-CAM analysis on the images from the test set for a better insight to the decision making process of the models and to make sure that the models are not overfitting on irrelevant features. Figure \ref{figGradcam} shows four successfully classified images from every class. AlexNet seems to focus more on consistency of the blood while VGG16 focuses on bands and distribution of the densities. For example, in spherocytosis, high density cells are more frequent which are successfully highlighted by AlexNet. Also in the healthy sample, the important areas are more uniformly distributed over the Percoll gradient bands.

\section{Conclusion}
We presented a novel hybrid approach based on fusion of Fourier transforms with convolutional neural networks for classification of Percoll gradients. We were able to show proof of concept to diagnose patients based on peripheral blood Percoll samples rather than genetic tests and without microscopy. This highly increases the applicability of the method in less developed and rural areas where access to the facilities is limited. The simplicity of this method once applied in practice can considerably cut the cost of diagnosis compared to conventional methods used for some of those blood disorders.
More experiments with bigger and more variant dataset, robustness against different illuminations, and clinical protocols are among the future steps of this study. Analysis of patient samples for assessment of the state of their disorder and the effectiveness of therapy is also another exciting topic for exploration. 

\subsection*{Compliance with Ethical Standards}
The protocol of the study complies with the World Medical Association Declaration of Helsinki, ICH-GCP guidelines and the local legally applicable requirements was approved by the local Ethics Committee.

\subsection*{Acknowledgment}
Special thanks to Prof. Ariel Koren and Dr. Carina Levin from the Emek Medical Center in Afula who made this work possible. This project has received funding from the European Union’s Horizon 2020 research and innovation programme under grant agreement No 675115 — RELEVANCE — H2020-MSCA-ITN-2015/ H2020-MSCA-ITN-2015. The work of L.L. was funded by UZH Foundation.
C.M. and A.S. have received funding from the European Research Council (ERC) under the European Union’s Horizon 2020 research and innovation programme (Grant agreement No. 866411). S.A. was supported by the PRIME programme of the German Academic Exchange Service (DAAD) with funds from the German Federal Ministry of Education and Research (BMBF) by the time of conducting this project.
\bibliographystyle{IEEEbib}
\bibliography{main}

\end{document}